\documentclass[letterpaper, 10 pt, conference]{ieeeconf}  % Comment this line out if you need a4paper

\IEEEoverridecommandlockouts                              % This command is only needed if 
\usepackage{amsmath}
\usepackage{amsthm}
\usepackage{algorithm}
\usepackage[noend]{algpseudocode}
\usepackage{amssymb}  % assumes amsmath package installed
\usepackage[dvipsnames]{xcolor}
\usepackage{booktabs}
\usepackage{hyperref}
\usepackage{mathtools}
\usepackage{multirow}
\usepackage{siunitx}
\usepackage{cite}
\usepackage{graphicx}
\usepackage{amsmath}
\usepackage{tikz}
\usetikzlibrary{arrows.meta, positioning}
\usepackage{comment}
\usepackage[caption=false,font=footnotesize]{subfig}

\newsavebox{\nnbox}
\newcommand{\Xs}[0]{\ensuremath{\mathcal{X}}}						% X
\newcommand{\Expect}{{\rm I\kern-.3em E}}				% expectation
\newcommand{\Prob}[1]{\text{P}\left(#1\right)}						% probability
\newtheorem{assumption}{Assumption}

\newenvironment{eqs*}{\begin{equation}\aligned}{\endaligned\end{equation}}
\newenvironment{eqs}[1]{\equation\label{#1}\aligned}{\endaligned\endequation}

\makeatletter
\renewenvironment{proof}[1][\proofname]{\par
\pushQED{\qed}%
\normalfont \topsep6\p@\@plus6\p@\relax
\trivlist
\item\relax
{\itshape
#1\@addpunct{.}}\hspace\labelsep\ignorespaces
}{%
\popQED\endtrivlist\@endpefalse
}
\makeatother
\graphicspath{{figures/}}

\algnewcommand\And{\textbf{and} }
\algnewcommand\Or{\textbf{or} }

\algrenewcommand\algorithmicforall{\textbf{parallel-for}}

\title{\LARGE \bf
Temporal Difference Learning of\\Probabilistic Reach-Avoid Value Functions
}

\author{Elias Fontanari$^{1}$, Pietro Noah Crestaz$^{1,2}$, Matteo Saveriano$^{1}$ and Andrea Del Prete$^{1}$%
\thanks{$^{1}$Industrial Engineering Department, University of Trento, Trento, Italy.}%
\thanks{$^{2}$LAAS-CNRS, Université de Toulouse, CNRS, Toulouse, France.}%
}

\begin{document}

\maketitle

\thispagestyle{empty}
\pagestyle{empty}

%%%%%%%%%%%%%%%%%%%%%%%%%%%%%%%%%%%%%%%%%%%%%%%%%%%%%%%%%%%%%%%%%%%%%%%%%%%%%%%%
\begin{abstract}
%To synthesize safe controllers and determine whether, from a given state, the task of interest can be successfully completed, it is often necessary to know the Reach-Avoid set, i.e., the set of states of a system from which the task can be completed without violating critical constraints. Robotic systems are typically deployed in the presence of noise and uncertainty. Classical methods for computing reach-avoid sets are based on Hamilton-Jacobi Reachability, which treats disturbances as an adversary interfering with the worst possible action from a bounded set. This can result in overly conservative sets in cases where the disturbance does not act adversarially.

%Addressing this drawback, we present a novel method for computing probabilistic reach-avoid sets, expressing the probability of safely completing the task and that can work even with disturbances whose support is unlimited. We compare it to HJR sets on systems of increasing complexity, up to HalfCheetah, showing that different, less conservative and more accurate sets are obtained for random disturbances.

Reach-avoid sets, defining the states from which a task can be completed while avoiding unsafe regions, are fundamental tools for deploying robots in environments subject to disturbances and uncertainty, as well as for runtime task-feasibility monitoring, allowing prediction of whether a state is safe and will lead to successful task completion. Classical Hamilton-Jacobi Reachability (HJR) and its recent neural extensions (e.g., Lipschitz Continuous Reachability Learning, DeepReach) treat disturbances adversarially, computing the reach-avoid set under worst-case disturbances from a bounded set. This guarantees safety but can yield overly conservative sets when disturbances are stochastic rather than adversarial, and additionally requires the disturbance set to be known or specified in advance.

We propose Probabilistic reach-avoid Value Function (PVF), a self-supervised method, applicable to black-box systems, for computing reach-avoid sets as the states with a certain reach-avoid probability under stochastic, potentially unbounded disturbances. We evaluate our approach against HJR-based baselines on systems of increasing complexity, up to a Hopper locomotion task, and show that our probabilistic sets are less conservative and more accurate than adversarial HJR sets when disturbances are non-adversarial.

\end{abstract}

%%%%%%%%%%%%%%%%%%%%%%%%%%%%%%%%%%%%%%%%%%%%%%%%%%%%%%%%%%%%%%%%%%%%%%%%%%%%%%%%

\section{Introduction}

\begin{figure*}
\centering
\includegraphics[width=\linewidth]{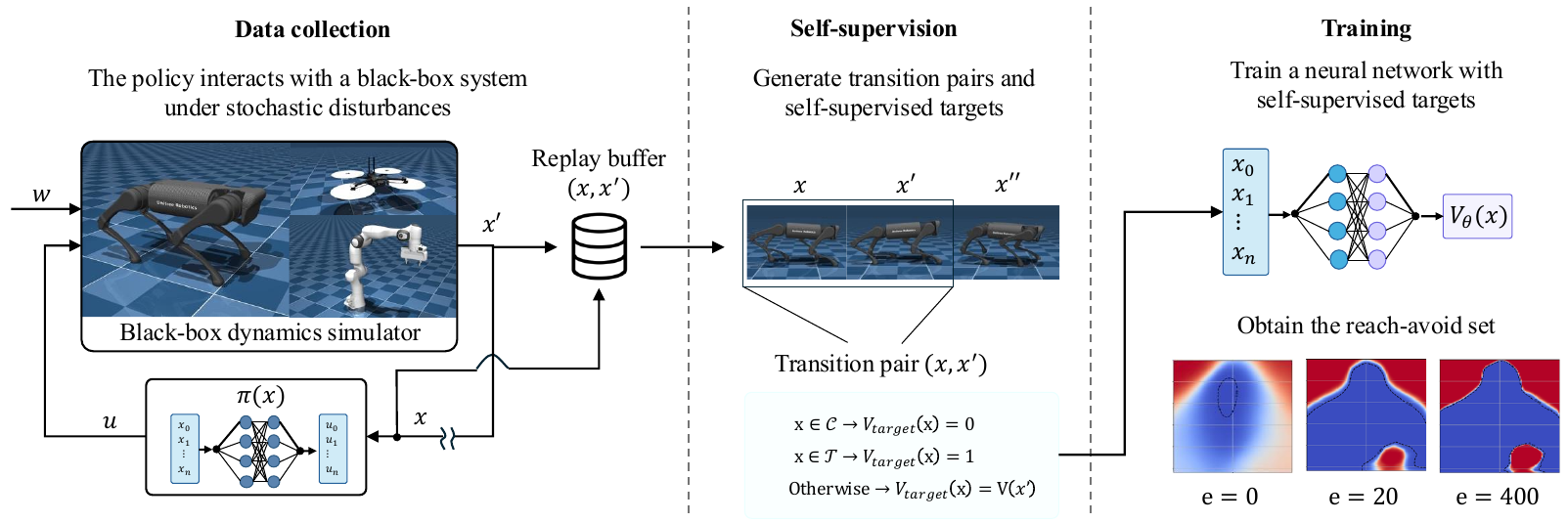}
\caption{Overview of the PVF pipeline. \textit{Data collection:} Trajectories are collected by running a nominal policy $\pi(x)$ on black-box systems (e.g., legged robots, manipulators, quadrotors) subject to stochastic disturbances $w$, and stored as transition pairs $(x, x')$ in a replay buffer. \textit{Self-supervision:} Piecewise target labels $V_{\text{target}}(x)$ are assigned based on constraint violation ($0$), task completion ($1$), or bootstrapped value evaluation ($V_\theta(x')$). \textit{Training:} A neural network parameterizing $V_\theta(x)$ is optimized via temporal-difference loss to output the continuous reach-avoid success probability.}
\label{fig:scheme}
\end{figure*}

Autonomous robotic systems are increasingly deployed to perform complex tasks in environments subjected to different sources of noise and uncertainty. Determining, from a given state, whether a task of interest can be completed while satisfying critical safety constraints can often be framed in terms of the \emph{Reach-Avoid set}: the set of states from which the system can reach a target region while avoiding an unsafe one. Avoid and Reach-Avoid sets are essential in runtime safety monitoring, as done in model predictive shielding \cite{Bastani2021}. Reach-avoid sets are also a useful tool for computing Control Invariant Sets (CIS) \cite{Blanchini1999}: if the target set used to compute a reach-avoid set is itself control invariant, the resulting reach-avoid set is also control invariant, and typically larger. CISs are essential tools in safe control methods such as Model Predictive Control (MPC) \cite{Grune2017}, since constraining the terminal state of the associated optimal control problem to lie in a CIS ensures recursive feasibility and constraint satisfaction.
Therefore, the accurate computation of reach-avoid sets is extremely useful for deploying robots in safety-critical settings.

\subsection{Related works}
% \ms{Forse 13 referenze sono un po' poche. Poi non vedo citato VBOC che è inerente. Non servirebbe un confronto perchè non considera rumore, dunamiche sotto attuate etc. Forse i lavori su receding (e altri approcci su recursive feasibility) potrebbero essere inseriti come modi di applicare i CIS?}   \EF{Ho aggiunto VBOC in fondo alla sezione related works. Aggiungere gli altri paper a me sembra un po' forzato, perchè si usano altre classi di set. Come li inserireste per non dare una sensazione di forzatura?}

Existing approaches to computing reach-avoid sets are predominantly grounded in Hamilton-Jacobi Reachability (HJR)~\cite{Bansal2017}, which treats disturbances as an adversary selecting the worst-case action from a bounded set. 
This comes with two main drawbacks: i) these methods cannot be applied to disturbances with unbounded support, such as a Gaussian distribution, and ii) when the disturbance is random they tend to be extremely conservative. 
Classic HJR methods rely on grid-based dynamic programming over the state space, which scales exponentially with the system's dimensionality, rendering them intractable for high-dimensional systems.

For this reason, the computation of reach-avoid sets has been addressed mainly by two lines of work, alleviating the curse of dimensionality: physics-informed neural networks such as DeepReach~\cite{Bansal2020}, and reinforcement learning~\cite{Fisac2019, Li2025}. 
DeepReach learns the reach-avoid set through a curriculum learning procedure that enforces the conditions of the associated Hamilton-Jacobi-Isaacs partial differential equation (PDE). 
RL-based methods, instead, learn a critic that encodes the safe set, replacing the PDE-based Hamilton-Jacobi formulation with a Bellman recursion that can be solved via RL algorithms. 
However, in the adversarial setting the disturbance policy must be learned jointly with the control policy, turning training into a min-max optimization problem that is notoriously unstable and slow to converge. 
In addition, the theoretical guarantees typically rely on regularity assumptions on the dynamics, most commonly Lipschitz continuity, which may not hold for hybrid or contact-rich systems.
DeepReach has been tested on large nonlinear systems such as quadrotors, but, unlike RL methods, it cannot be applied to non-differentiable dynamical systems, such as legged robots or black-box systems, since the dynamics is needed in the computation of the Hamiltonian.

A separate line of work, stochastic reachability analysis~\cite{Abate2008}, \cite{Summers2010}, addresses reach-avoid problems directly as probabilities of success under a stochastic disturbance model, formulated via dynamic programming over the system's transition kernel in a measure-theoretic setting. 
While closer in spirit to our probabilistic formulation, these approaches rely on the numerical solution of the associated Bellman recursion over a gridded or otherwise explicitly represented state space, which precludes their use with the high-dimensional, black-box dynamics typical of robotic simulators.

In another line of work~\cite{Srinivasan2020} the authors used actor-critic RL to learn a safe policy, where the critic expresses a pseudo-probability of future failure. 
% , due to the discount applied to future violations. 
However, several aspects of this formulation limit its use as a set-computation method. 
First, the critic addresses only the avoid component: it estimates the probability of failure with no notion of task completion and cannot distinguish states from which the task is achievable from states that are merely never-failing. 
Second, the critic is defined with respect to the policy induced by the critic itself, a self-referential construction that the authors report to be excessively pessimistic early in training and that requires implementation-level corrections, namely computing Bellman targets under the unconstrained policy and maintaining a separate near-on-policy replay buffer. 

Another method capable of computing reach-avoid sets, specifically used to compute CISs, is Viability Boundary Optimal Control (VBOC) \cite{LaRocca2023}, which avoids resorting to HJR by directly identifying states on the boundary of the reach-avoid set and training a neural network to represent it. The target set corresponds to the system's equilibria, while the avoid set represents collisions with the environment or joint limit violations. By directly learning the boundary of the reach-avoid set, VBOC achieves higher data efficiency compared to HJR, but its main drawbacks are that it does not account for disturbances and can only handle fully-actuated systems.

\subsection{Contribution}
We introduce \textbf{PVF} (\textbf{P}robabilistic Reach-Avoid \textbf{V}alue \textbf{F}unctions), addressing the typical drawbacks of HJR methods in settings where there is no adversary, but simply non-adversarial stochastic noise, as is often the case in robotics scenarios, wind disturbances acting on an aerial vehicle being a representative example.
By learning a value function that expresses, for each state, the probability of reaching the target set without ever entering the failure set, we are able to assess safe target reaching in a graded fashion: states that succeed under most disturbance realizations, and fail only under a highly unfavourable sequence of disturbances, are distinguished from states that fail under typical conditions.

Crucially, the entire procedure treats the system as a black box: it requires only the ability to sample transitions from the environment, with no access to an analytical model, no differentiability of the dynamics, and no knowledge of the disturbance distribution. 
This makes the approach directly applicable to contact-rich and hybrid systems, such as legged robots, for which a closed-form model is either unavailable or non-differentiable, and where PDE-based methods cannot be applied. Fig.~\ref{fig:scheme} provides an overview of the proposed method.

Additionally, our value function provides interpretable probabilities, as opposed to those computed by HJR methods, which provide either the worst-case constraint violation the trajectory will incur, or the value of the target-set membership function at the time the target is reached.
% and it does not require the failure and target sets to be expressed analytically (e.g., via a level-set equation), but simply through an indicator function denoting membership in each set.

Summarizing, the contributions are the following: (i) a novel self-supervised method for computing probabilistic reach-avoid sets under stochastic disturbances with potentially unbounded support, without assuming worst-case behavior; (ii) a comparison against HJR-based baselines on systems of increasing complexity, up to a Hopper locomotion task, showing that the resulting sets are more interpretable, less conservative and more accurate under non-adversarial disturbances.

\section{Learning the Probabilistic Value Function (PVF)}

\subsection{Problem formulation}
We consider stochastic discrete-time dynamical systems of this form: 
\begin{equation}
    \begin{aligned}
        &x' = f(x, u, w), \\
        &x \in \mathcal{X} \subseteq \mathbb{R}^n,\ u \in \mathcal{U} \subseteq \mathbb{R}^m,\ w \sim p_\mathcal{W},
    \end{aligned}
\end{equation}
where $x$ is the current state, $x'$ the state reached applying the control $u$, and the stochastic disturbance $w$, drawn from a possibly unbounded distribution $p_{\mathcal{W}}$, to the system. The transition function $f: \mathcal{X} \times \mathcal{U} \times \mathcal{W} \rightarrow \mathcal{X}$ is treated as a \emph{black box}: it is unknown in closed form and can only be queried through a simulator or real-world rollouts, without access to its analytical form, gradients, or the explicit distribution $p_{\mathcal{W}}$.

Given a fixed (possibly stochastic) policy $\pi(\cdot): \mathcal{X} \rightarrow \mathcal{U}$, the objective is to compute, starting from each state $x$, the probability of reaching a target set $\mathcal{T}$ without passing through a constraint set $\mathcal{C}$, 
% \notin \mathcal{C}$ 
% (the probability is trivially zero if $x \in \mathcal{C}$) \ms{shall we assume already that $x \notin \mathcal{C}$}
i.e. the reach-avoid probability:
\begin{eqs}{}
    V(x) \triangleq \mathbb{P}\!\left(\tau_{\mathcal{T}} < \tau_{\mathcal{C}} \mid X_0=x\right),
\end{eqs}
where $\tau_{\mathcal{T}}$ and $\tau_{\mathcal{C}}$ denote the first hitting times of the target and constraint sets, respectively.
Given this probability function, we can define probabilistic reach-avoid sets, for a given probability threshold, as the superlevel set of the function.

Our approach does not require an explicit expression for the target and constraint sets; it suffices to have a method that checks whether a given state belongs to them. 
To this end, we define two indicator functions, $\mathcal{I}_{\mathcal{T}}$ and $\mathcal{I}_{\mathcal{C}}$, associated with the target set $\mathcal{T}$ and the constraint set $\mathcal{C}$, respectively:
\begin{align}
    \mathcal{I}_{\mathcal{T}}(x) &=
    \begin{cases}
        1 & x \in \mathcal{T}, \,\, \text{target reached}\\
        0 & \text{otherwise}
    \end{cases} \\
    \mathcal{I}_{\mathcal{C}}(x) &=
    \begin{cases}
        1 & x \in \mathcal{C}, \,\, \text{constraints violated} \\
        0 & \text{otherwise}
    \end{cases}
\end{align}
The indicator functions can be negated with an overline bar, such as $\overline{\mathcal{I}}$.

\subsection{Method}
A state can be either in the target set, in the constraint set, or in neither of those. 
Assuming without loss of generality that target and constraint set have empty intersection\footnote{If that is not the case we can always redefine the target set by removing its intersection with the constraint set.}, these three cases are mutually exclusive. 
Let us now define the following three symbols to represent these three events, where $x_i$ is the state at the step $i$ in an episode starting from step $0$:
\begin{eqs}{}
    T_i:\quad & x_i \in \mathcal{T} \\
    C_i:\quad & x_i \in \mathcal{C} \\
    N_i:\quad & x_i \notin \mathcal{T} \land x_i \notin \mathcal{C}
\end{eqs}
Starting from a given state at timestep $t_0=0$ we define the event $S_0$ of eventually reaching the target set before violating any constraint:
\begin{eqs}{}
% V(x_0) = \Prob{ \exists t \ge 0: \mathcal{I}_{\mathcal{T}}(x_t) = 1 \wedge [\text{sup}_{i\in[0, t]} \mathcal{I}_{\mathcal{C}}(x_i)] = 0}
S_0 &= \bigvee_{t \ge 0} \left[ T_t \land \left( \bigwedge_{i\in[0, t-1]} N_i \right) \right] \\
&= T_0 \lor (N_0 \land T_1) \lor (N_0 \land N_1 \land T_2) \lor \dots
\end{eqs}
Since the event $N_0$ is common to all the terms after the first one, we can collect it to express this event recursively:
\begin{eqs}{}
S_0 &= T_0 \lor [N_0 \land \overbrace{(T_1 \lor (N_1 \land T_2) \lor \dots)}^{S_1}] \\
&= T_0 \lor (N_0 \land S_1)
\end{eqs}

% \EF{ riscriverei l'equazione così perchè altrimenti se si sviluppasse non verrebbe la 6 e N deve essere minore di 1 rispetto a T
% \begin{eqs}{}
% % V(x_0) = \Prob{ \exists t \ge 0: \mathcal{I}_{\mathcal{T}}(x_t) = 1 \wedge [\text{sup}_{i\in[0, t]} \mathcal{I}_{\mathcal{C}}(x_i)] = 0}
% S_0 &= T_0  \bigvee_{t \ge 1} \left[ T_t \land \left( \bigwedge_{i\in[0, t-1]} N_i \right) \right] \\
% &= T_0 \lor (N_0 \land T_1) \lor (N_0 \land N1 \land T_2) \lor \dots
% \end{eqs}}
% Since the event $N_0$ is common to all the terms after the first one, we can collect it to express this event recursively:
% \begin{eqs}{}
% S_0 &= T_0 \lor [N_0 \land \overbrace{(T_1 \lor (N_1 \land T_2) \lor \dots)}^{S_1}] \\
% &= T_0 \lor (N_0 \land S_1)
% \end{eqs}}
Exploiting the fact that $T_0$ and $N_0$ are mutually exclusive, we can translate this expression in terms of probabilities as:
\begin{eqs}{}
\Prob{S_0} &= \Prob{T_0} + \Prob{N_0 \land S_1} \\
&= \Prob{x_0 \in \mathcal{T}} + \Prob{x_0 \notin \mathcal{C} \land x_0 \notin \mathcal{T} \land S_1}
\end{eqs}
After the state $x_0$ is observed, the above probabilities on $x_0$ become binary and can be expressed using the indicator functions defined above:
\begin{eqs}{eq:reach_avoid_p}
\Prob{S_0 | X_0=x_0} &= \mathcal{I}_{\mathcal{T}}(x_0) + \overline{\mathcal{I}_{\mathcal{C}}}(x_0) \, \overline{\mathcal{I}_{\mathcal{T}}}(x_0) \, \Prob{S_1 | X_0=x_0}
%&= (1-\mathcal{I}_{\mathcal{C}}(x_0)) \max (\mathcal{I}_{\mathcal{T}}(x_0), \Prob{S_1})
\end{eqs}
This expression is our starting point for defining a recursive evaluation of the reach-avoid probability, which we want to learn as a Value function $V(x)$.
Based on~\eqref{eq:reach_avoid_p} we write the following expression for the Reach-Avoid value function:
% \begin{equation} \label{eq:bellman_reach_avoid}
%     V(x) = \bigl(1 - \mathcal{I}_{\mathcal{C}}(x)\bigr)\left(\max\left(\mathcal{I}_\mathcal{T}(x),\ 
%     \Expect[V(x')] \right) \right)
% \end{equation}
\begin{equation} \label{eq:bellman_reach_avoid}
    V(x_0) = \mathcal{I}_{\mathcal{T}}(x_0) + \overline{\mathcal{I}_{\mathcal{C}}}(x_0) \, 
    \overline{\mathcal{I}_{\mathcal{T}}}(x_0) \,
    \Expect[V(x_1) | X_0 = x_0]
\end{equation}
Evaluating the expectation of $V(\cdot)$ directly is intractable because the transition probability distribution $\Prob{x'|x, u}$ is unknown. However, given access to a simulator, it is possible to sample from the distribution and approximate this expectation without explicit knowledge of the transition dynamics using Temporal-Difference (TD) learning \cite{td_sutton_1988}:
\begin{equation} \label{eq:td_update}
    V(x) \leftarrow V(x) + \alpha\bigl(V_\mathrm{target}(x, x') - V(x)\bigr),
\end{equation}
where $\alpha \in [0, 1]$ is the user-defined learning rate. 
The target can be computed by evaluating~\eqref{eq:bellman_reach_avoid} in the three cases where $x$ belongs to $\mathcal{C}$, to $\mathcal{T}$, or to neither of those:
\begin{equation}
\label{eq:target}
    V_\mathrm{target}(x, x') :=
    \begin{cases}
        0 & \text{if } x \in \mathcal{C} \\
        1 & \text{if } x \in \mathcal{T} \\
        V(x') & \text{otherwise}
    \end{cases}
\end{equation}
As typically done in TD learning, the expectation of $V(\cdot)$ over the stochastic system dynamics in~\eqref{eq:bellman_reach_avoid} is replaced by the current estimate of $V(\cdot)$ for the observed next state $x'$.

\begin{assumption} \label{ass:almost_sure_hit}
Let us define the time at which the state enters either the target or the constraint set for the first time: $\tau(x)=\inf\{t\ge0: x_t \in \mathcal T \cup \mathcal C, x_0=x\}$. 
We assume that eventually the system reaches $\mathcal{T} \cup \mathcal C$ almost surely, that is: \(P(\tau(x) < \infty) = 1\) for every \(x \in \Xs\).
\end{assumption}

This assumption rules out closed regions in which the system can remain indefinitely without reaching either $\mathcal{T}$ or $\mathcal{C}$. In the presence of such regions, the undiscounted Bellman equation may admit multiple fixed points because no terminal boundary value is propagated to them.

% Un der the above assumption, the reach-avoid probability is the unique bounded solution of the stated Bellman equation with boundary values \(V=1\) on \(\mathcal T\) and \(V=0\) on \(\mathcal C\).

% \pnc{I would rewrite Equations (10)–(11) to consistently evaluate the next state $x'$ for clarity and reformulate as follows:
% \begin{equation}
%     V(x) \leftarrow V(x) + \alpha \bigl( V_{\text{target}}(x') - V(x) \bigr)
% \end{equation}
% where $V_{\text{target}}(x') = (1-\mathcal{I}_{\mathcal{C}}(x')) \max\bigl(\mathcal{I}_{\mathcal{T}}(x'), V(x')\bigr)$, which translates piecewise to:
% \begin{equation}
%     V_{\text{target}}(x') :=
%     \begin{cases}
%         0 & \text{if } x' \in \mathcal{C} \\
%         1 & \text{if } x' \in \mathcal{T} \\
%         V(x') & \text{otherwise}
%     \end{cases}
% \end{equation}
% } \EF{To me it seems more correct keeping x as argument, than if one substitutes x' the equation you wrote is recovered} \ms{I would definitely define $V_{\text{target}}$ in terms of the indicator functions (as Pietro suggested). Then, I slightly prefer (13) instead of (11).}

By using a neural network as a function approximator, the TD update~\eqref{eq:td_update} can be extended to the learning of the continuous reach-avoid probability. 
The loss function to be minimized is
\begin{equation}
    \min_{\theta} \sum_{(x,x') \in \mathcal{P}} \left\| V_\theta(x) - V_\mathrm{target}(x, x') \right\|^2
\end{equation}
where $\mathcal{P}$ is the set of the stored pair transitions, obtained by rolling out the system using the given policy~$\pi$.

\subsection{Algorithmic implementation}
To train a value function, the procedure is reported in Algorithm \ref{alg:pvf-training}.
First, trajectories are generated in a data collection phase by sampling random initial states and executing the given policy. Subsequently, transition pairs are extracted from the trajectories and inserted into the paired dataset $\mathcal{P}$. Finally, the training phase is executed. Each pair is used to compute the target for $V_{\theta}(x)$ as described in Eq.~\ref{eq:target}, and the network learns the reach-avoid probability by minimizing the mean squared error with respect to the target. The training is stopped when the change of the network parameters $\theta$ is small and lower than a chosen threshold $\epsilon$. We noticed that introducing another network to compute the target when needed, improves the stability of the learning process. After each optimization step, the target network parameters $\theta_{\text{target}}$ are updated using Polyak averaging \cite{lillicrap2019}, with small soft update coefficient $\tau_\theta$, in the order of $0.001$. 
%\begin{enumerate}
 %   \item starting from $n$ random state, rollout the system with a given policy, interrupting the rollout if a constraint is violated or if the target is reached and saving the trajectories
  %  \item from the saved trajectories extract the transition pairs
   % \item train the network until the relative change of the parameters $\theta$ are negligible 
%\end{enumerate}

While our theory discusses an infinite-horizon hitting probability, the algorithm uses a finite episode length. 
However, this timeout is only a data-collection cutoff for an infinite-horizon quantity because the value targets are computed based on the current estimate of the reach-avoid probability.
Therefore, as the number of training epochs grows to infinity, so does the horizon considered by the computed value function.

\begin{algorithm}[t]
\caption{Probabilistic Value Function (PVF) Training}
\label{alg:pvf-training}
\begin{algorithmic}[1]
\Require Simulator $\mathcal{S}$, control policy $\pi$, number of episodes $n_{\mathrm{episodes}}$, network parameters $\theta$, target network parameters $\theta_{target}$, soft update coefficient $\tau_\theta$, stopping threshold $\epsilon$, maximum number of epochs $E_{\max}$
\Ensure Trained parameters $\theta$

\State $\mathcal{D} \gets \emptyset$ \Comment{trajectory buffer}
\For{$i = 1$ to $n_{\mathrm{episodes}}$}
    \State Sample initial state $x_0 \sim \mathcal{X}$
    \State $\tau \gets \text{Rollout}(\mathcal{S}, \pi, x_0)$, stopping early if $\mathcal{C}$ or $\mathcal{T}$ are 
    \Statex \hspace{\algorithmicindent}%
    reached
    \State $\mathcal{D} \gets \mathcal{D} \cup \{\tau\}$
\EndFor

\State $\mathcal{P} \gets$ ExtractTransitionPairs$(\mathcal{D})$

\For{$e = 1$ to $E_{\max}$}
    \State $\theta_{\text{old}} \gets \theta$
    \State Randomly split $\mathcal{P}$ in minibatches $B_i$, 
    \Statex \hspace{\algorithmicindent}%
    creating batch set $\mathcal{B} = \{B_1,...,B_n\}$
    \For{each batch $B_i \in \mathcal{B}$ }
    %\State Sample minibatch $\mathcal{B}\subset\mathcal{P}$
    \For{each pair $(x, x') \in B_i$}
        \If{$x \in \mathcal{C}$}
            \State $y \gets 0$
        \ElsIf{$x \in \mathcal{T}$}
            \State $y \gets 1$
        \Else
            \State $y \gets V_{\theta_{\text{target}}}(x')$
        \EndIf
        \State Label pair $(x, x')$ with target $y$
    \EndFor
    \State Update $\theta$: $\min\limits_{\theta} \sum_{(x,x') \in B_i} \|V_\theta(x) - y\|^2$
    \State Update target network:
    \Statex \hspace{1.1cm}%
    $\theta_{\text{target}} \leftarrow \tau_\theta \, \theta + (1 - \tau_\theta) \, \theta_{\text{target}}$
    % \State $\text{abs\_norm} \gets \|\theta - \theta_{\text{old}}\|$
    \EndFor
    \State $\text{rel\_norm} \gets \|\theta - \theta_{\text{old}}\| \, / \, \|\theta_{\text{old}}\|$
    \If{$\text{rel\_norm} < \epsilon$}
        \State \textbf{break}
    \EndIf
\EndFor

\State \Return $\theta$
\end{algorithmic}
\end{algorithm}

\subsection{Comparison with related methods} \label{ssec:comparison}
We compare our approach with two representative learning-based safety methods: Lipschitz Continuous Reachability Learning (LCRL)~\cite{Li2025}, which learns robust reach-avoid sets, and Safety Q-functions for Reinforcement Learning (SQRL)~\cite{Srinivasan2020}, which estimate the future failure risk of a safety-constrained policy.
Both methods additionally synthesize or adapt a controller. 
In contrast, our method focuses on estimating the reach-avoid probability of a given closed-loop system. 
We therefore restrict the following discussion to the value functions learned by the three approaches.

LCRL belongs to the family of Hamilton--Jacobi reachability methods and adopts a robust formulation: its value function determines whether the target can be reached safely for every disturbance realization within a prespecified bounded set. 
Our method instead models disturbances stochastically and estimates the probability of reaching the target before entering the constraint set. 
Consequently, the two value functions encode different notions of feasibility. LCRL provides a worst-case, binary characterization through the sign of its value function, whereas our value function quantifies the likelihood of task success under the disturbance distribution. 
The latter can distinguish states having different success probabilities even when neither state admits a worst-case guarantee. 
Moreover, our formulation does not require a bounded disturbance set and therefore also accommodates distributions with unbounded support. To enable a direct empirical comparison, we consider bounded stochastic disturbances, for which LCRL can be applied by treating every disturbance in the support as adversarial. 
Such a comparison highlights the conservatism introduced when disturbance likelihoods are discarded.
SQRL is closer to our approach because its safety critic is also defined through an expectation over stochastic transitions. 
Its objective, however, differs from reach-avoid feasibility. 
SQRL estimates, for each state--action pair, the discounted risk of eventually entering a failure set, while task completion is handled separately through the policy reward. 
It therefore does not include an explicit target set in the safety-critic formulation. 

This distinction is particularly relevant over long or infinite horizons. 
If a system remains exposed to a nonzero failure risk indefinitely, its eventual failure probability may equal one over a large portion of the state space, providing little information about task feasibility. 
SQRL mitigates this issue through a finite horizon and a safety discount factor $\gamma_{\mathrm{safe}}<1$, which emphasize near-term failures but cause the resulting critic to represent discounted risk rather than an undiscounted hitting probability. 
Our target set instead defines successful termination explicitly. 
Together with the constraint set, it supplies the two boundary conditions needed to distinguish successful and unsuccessful task completion. 
Under the standard Assumption~\ref{ass:almost_sure_hit}, these boundary conditions uniquely determine the reach-avoid probability and provide terminal supervision for TD learning. 
Thus, the two approaches are complementary: SQRL is designed to constrain failure risk during policy learning, whereas our method is designed to quantify the probability that a specified task can be completed safely.

\section{Results}
To assess the accuracy and characteristics of our method\footnote{Code will be made available upon acceptance}, we compared it against the two above-mentioned state-of-the-art methods: LCRL~\cite{Li2025}, and SQRL~\cite{Srinivasan2020} without discounting.

\subsection{Introductory example: Toy problem}
% To illustrate the difference between the proposed method and HJR methods, we first consider a simple scalar toy problem, taken from \cite{Li2025}. The dynamics, target and avoid sets are defined as: 
% \begin{align*}
%     &x' = 1.01 \; x + 0.01 \; (u + w)\\ 
%     &\mathcal{T} = \{x \;| \; -2 < x \le -1\} \\
%     &\mathcal{C} = \{x \;| \; x \le -2\}
% \end{align*}
% Assuming the control satisfies $|u|\le1$ and the disturbance satisfies $|w|\le4$, as in the case of Gaussian noise truncated at $4\sigma$ with $\sigma = 1$, the value functions obtained with our method and with LCRL~\cite{Li2025} are depicted in Fig.~\ref{fig:toy}. To generate the dataset we have used the control policy computed by  LCRL. \pnc{Simple problem, but I would still formulate it a bit more rigorusly, stating that x is the state variable, u the control input etc... and constraints can be put together into the same formulation. Suggested reformulation: }

\begin{figure}
    \centering
    \includegraphics[width=\columnwidth]{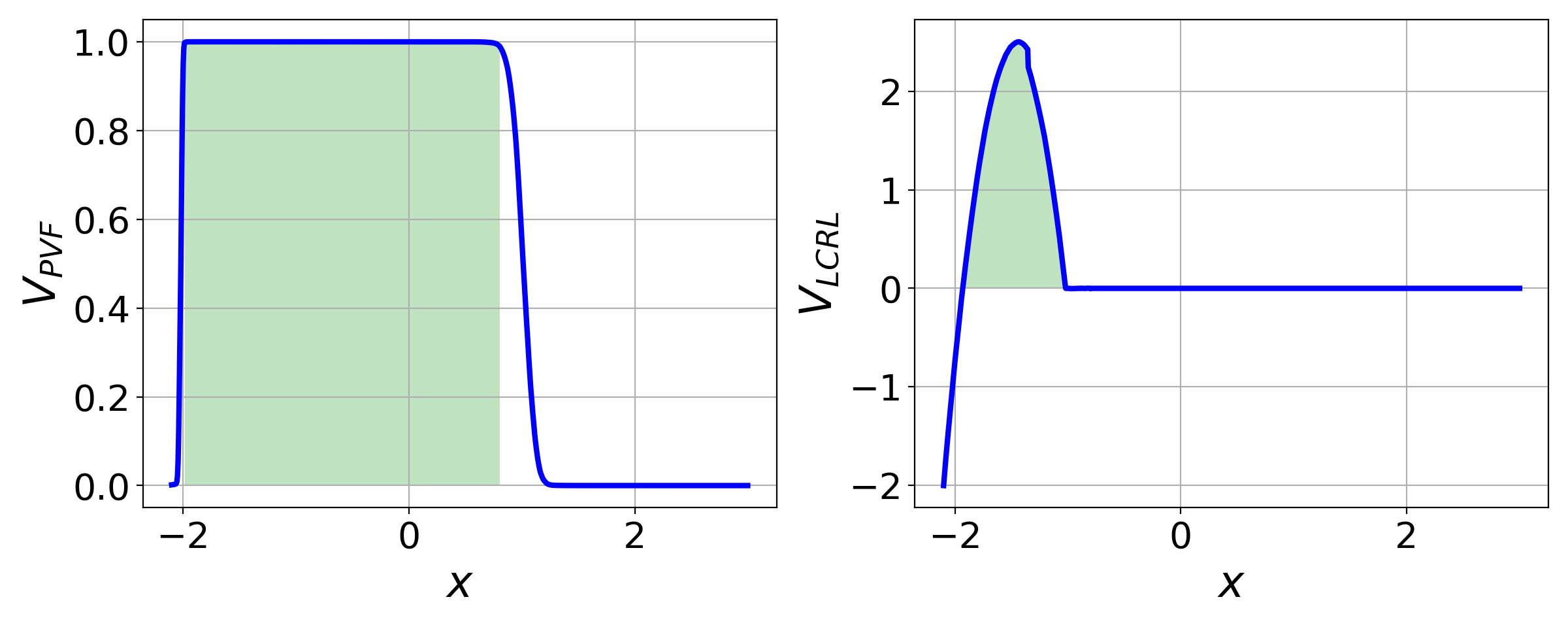}
    \caption{{Reach-avoid value functions on the scalar toy problem.} Comparison between the proposed Probabilistic Value Function ($V_{\text{PVF}}$, left) and the adversarial Lipschitz Continuous Reachability Learning baseline ($V_{\text{LCRL}}$, right). Green regions indicate states identified as safe (i.e., able to reach the target set without violating constraints).}
    \label{fig:toy}
\end{figure}

To illustrate the difference between the proposed method and HJR methods, we first consider a simple scalar toy problem, taken from \cite{Li2025}. Let $x \in \mathcal{X} \subseteq \mathbb{R}$ denote the scalar state, $u \in \mathcal{U} \subset \mathbb{R}$ the control input, and $w \in \mathcal{W} \subset \mathbb{R}$ the stochastic disturbance. The dynamics, target set $\mathcal{T}$, and constraint set $\mathcal{C}$ are defined as:
\begin{equation}
\begin{aligned}
    x' &= 1.01 \, x + 0.01 \, (u + w), \\
    \mathcal{T} &= \{ x \in \mathbb{R} \mid -2 < x \le -1 \}, \\
    \mathcal{C} &= \{ x \in \mathbb{R} \mid x \le -2 \}, \\
    \text{s.t.} \quad & |u| \le 1, \quad |w| \le 4,
\end{aligned}
\label{eq:toy_problem}
\end{equation}
where the disturbance bound $|w| \le 4$ corresponds to a zero-mean Gaussian distribution with $\sigma = 1$ truncated at $4\sigma$. To generate the dataset $\mathcal{D}$ for evaluating our method, we collect transitions under the control policy $\pi(x)$ computed by LCRL~\cite{Li2025}. The resulting value functions obtained via our method ($V_{\text{PVF}}$) and LCRL ($V_{\text{LCRL}}$) are reported in Fig.~\ref{fig:toy}.

With this level of noise, if the worst-case disturbance is always chosen, the system can never reach the target (except when it is already inside the target), as shown in the right panel of Fig.~\ref{fig:toy}. 
In contrast, our method marks a larger portion of the state space as successful (i.e., states from which the target can be reached with high probability). 
This example illustrates that our method is substantially different from reach-avoid HJR, and that in situations where the noise is not adversarial, it can be significantly less conservative.

\begin{figure*}[t]
    \centering
    \includegraphics[width=\textwidth]{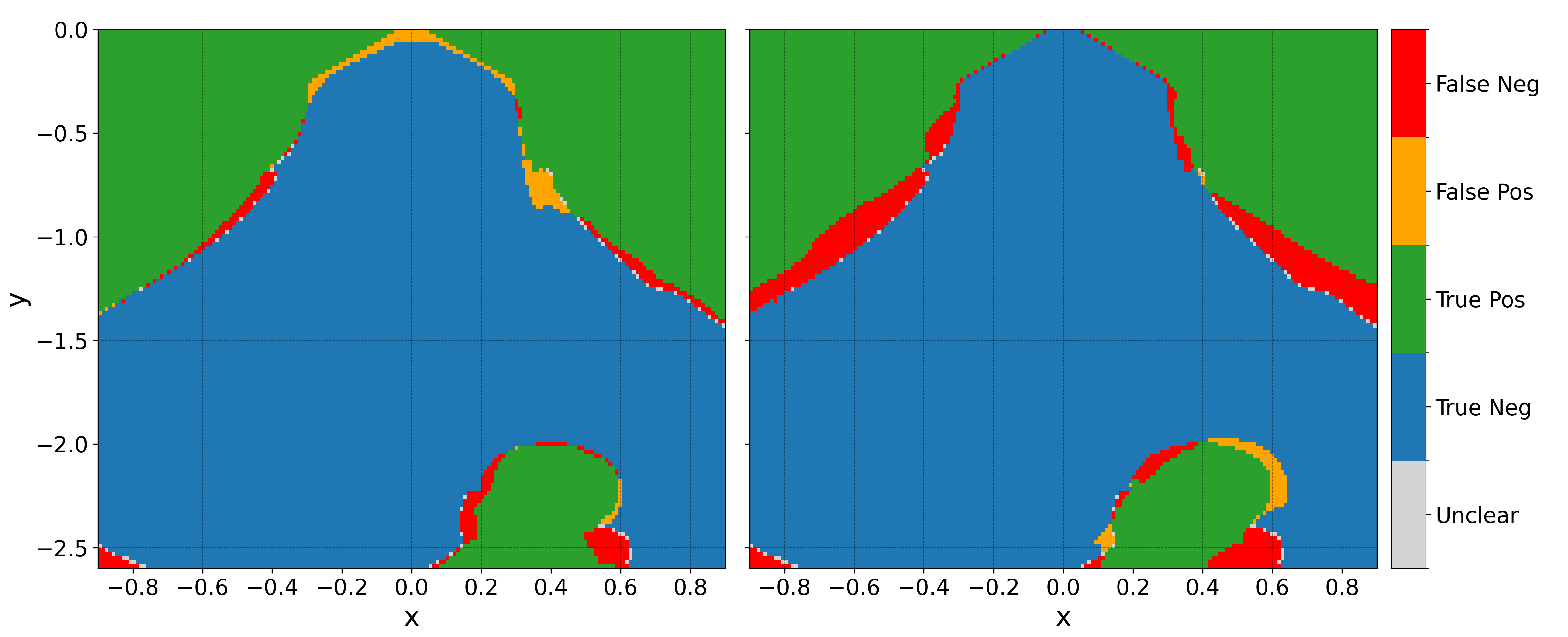}
    \caption{{Visual comparison between PVF (left) and LCRL (right) in terms of accuracy, for Drones racing experiment.} $x$ and $y$ indicate the initial position of the ego drone in the horizontal plane. The opposing drone starts at $x=0.4$ m and $y=0.0$ m }
    \label{fig:Visual comparison between PVF and LCRL prediction}
\end{figure*}
\subsection{Value function accuracy: LCRL comparison}
We assessed the accuracy of our method on two other systems of increasing difficulty: Drones Racing \cite{Li2025} and Hopper \cite{mujoco_playground_2025,Durrant-White2012}. 
Since PVF requires a policy, for a fair comparison we always tested all methods with the same policy. 
% For the Drone Racing test, we used the policy obtained via LCRL, while for Hopper and Half Cheetah, for which LCRL struggled to find a good control policy, we supplied LCRL with two locomotion policies obtained via classic RL, leaving LCRL the task of learning the worst-case disturbance and the reach-avoid set.
For the Drone Racing experiment, we used the policy learned directly via LCRL \cite{Li2025}. For Hopper, where LCRL struggled to optimize a control policy, we instead provided LCRL with a pre-trained locomotion policy obtained using the standard RL algorithm Proximal Policy Optimization (PPO) \cite{schulman2017ppo}. This isolates LCRL's role to learning the worst-case disturbance and the corresponding reach-avoid set under a fixed policy. The PVF network was a 5-layer MLP with $256$ hidden units per layer and GELU activations, with a linear (unactivated) scalar output. LCRL's actor (present in Drones Racing only), disturbance, and critic networks used a 4-layer, 512-unit MLP with ReLU activations, taking the state (plus action, for the critic) as input, with role-dependent outputs: a scalar estimate for the critic, and a bounded control for the actor/disturbance networks.

In all tests, LCRL was run until the critic loss stabilized below a chosen threshold ($10^{-4}$ for Drones Racing and $10^{-6}$ for Hopper), or until the maximum number of epochs ($160$) was reached.

To obtain a statistically grounded estimate of the true reach-avoid probability of the policy $\pi$ at a given state $x$, we treat each closed-loop rollout as an independent Bernoulli trial and estimate the reach-avoid probability $p(x)$ via Monte Carlo sampling with an adaptive stopping rule based on the Wilson score confidence interval.
This is known to give substantially better coverage than the naive normal-approximation (Wald) interval, especially for small sample sizes or proportions near $0$ or $1$~\cite{brown2001interval}.

For each sampled state $x_i$, we run a fixed budget of $n = 4000$ rollouts and
form the estimate $\hat p_i = s_i/n$ from the observed successes $s_i$, together
with its Wilson interval $[p_{lo,i},\,p_{hi,i}]$ at confidence level
$1-\beta = 99.5\%$. The state is classified \emph{truly successful} if
$p_{lo,i} \geq \ell$, \emph{truly unsuccessful} if $p_{hi,i} < \ell$, and
\emph{unclear} if the interval straddles the target safety level $\ell = 0.99$.
This provides the Monte Carlo reference against which the learned reach-avoid
classifications are evaluated.
The ground-truth probability level 
$\ell=0.99$ was chosen as a practical trade-off: it is high enough to clearly discriminate unsuccesful states, while remaining statistically tractable, since higher levels require substantially more rollouts to certify at the confidence chosen. Certifying a success probability of exactly 
$1$, by contrast, is fundamentally problematic, as no finite number of trials can rule out the possibility that an unfavorable event has simply not yet been observed.
The value functions approximations $V_{PVF}(x)$ and $V_{LCRL}$ are scored against this ground truth certificates over the clear states: a \emph{true positive} (TP) is a truly unsuccessful state that $V$ also predicts unsuccessful, a \emph{false negative} (FN) a truly unsuccessful state that $V$ predicts successful, a \emph{false positive} (FP) a truly successful state that $V$ predicts unsuccessful, and a \emph{true negative} (TN) a truly successful state that $V$ predicts successful. For each system, these metrics were evaluated on $1000$ states reached after a random number of policy steps from randomly sampled initial states, discarding those that ended inside the target or the constraint set. With $n = 4000$ rollouts per state, the fraction of unclear cases remained
below $4.5\%$ across all experiments.

\subsubsection{Drones racing}  

This scenario, adapted from the LCRL paper~\cite{Li2025}, formulates a drone take-over maneuver as a reach-avoid problem involving two quadrotors racing through a gate, where the ego drone must overtake safely its opponent. The opposing drone heads straight for the gate under a fixed LQR controller, but its acceleration is not perfectly known to the ego drone; this uncertainty enters the model as a disturbance bounded in norm by $2~\text{m/s}^2$. The noise distribution is assumed to be a zero-mean Gaussian truncated at $3\sigma$, with $\sigma=\frac{2}{3}$$\text{m/s}^2$. Our goal is to compute the probability that the ego drone can safely overtake the opponent drone and reach the gate first, regardless of this disturbance.

The joint system state is 12-dimensional, with each drone described by its position and velocity along the three spatial axes. Along each axis, the drones follow double-integrator dynamics, driven by acceleration inputs. While the dynamics themselves are linear, the problem is nonlinear due to the constraints defining the avoid set. In particular, the safety constraint requires the ego drone to stay outside the airflow generated by the other drone, combining a squared norm with a max operator.% so that the required separation grows with their height difference.

\begin{figure}
    \centering
    \includegraphics[width=\columnwidth]{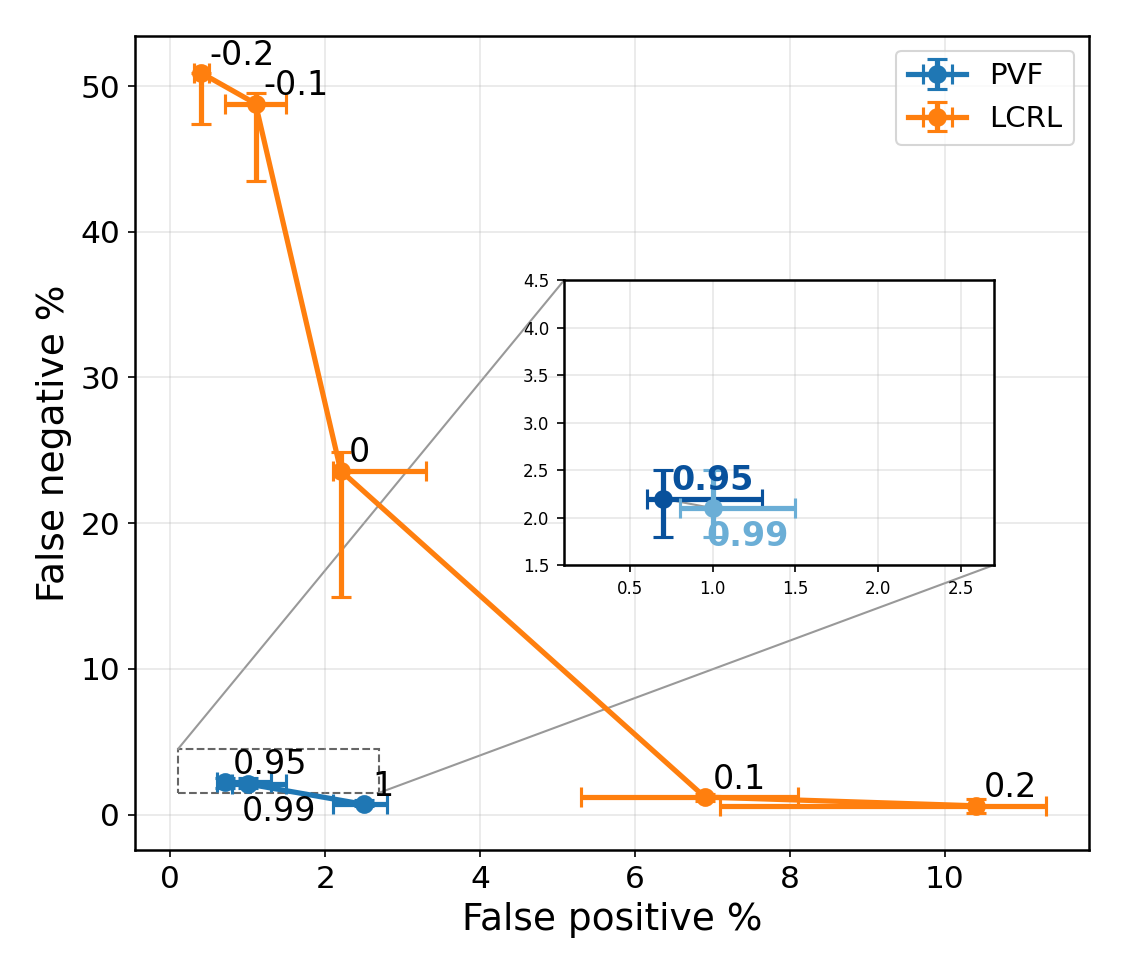}
    \caption{{Comparison between PVF and LCRL in terms of accuracy, for Drones racing experiment.} Thresholds for fail detection reported near the medians.}
    \label{fig:drones}
\end{figure}

The results of three full comparisons under different seeds are presented in Fig.~\ref{fig:drones}, where the FP and FN of each method are reported, showing minimum, maximum and median errors across runs. We also report the classification threshold used by PVF and LCRL.
At the nominal thresholds ($0.99$ for PVF and $0$ for LCRL), PVF performed remarkably better in terms of FN and moderately better in terms of FP. The number of FN, the most dangerous classification failure due to a missed alarm, was consistently smaller for PVF, except when the LCRL threshold was increased over $0.1$, which however increased the FP (i.e., false alarms) above $6\%$. Fig.~\ref{fig:Visual comparison between PVF and LCRL prediction} shows a visual comparison between the reach-avoid sets computed by PVF and LCRL, tested against Wilson confidence intervals, on a particular section of the state space, as in~\cite{Li2025}, by varying only the $xy$ position of the ego drone in the horizontal plane. This figure highlights the higher number of FN for LCRL, particularly for $y> - 1.5$ m.

% \begin{figure}[h]
% \centering
% %\begin{subfigure}[b]{0.235\textwidth}
% \subfloat[]{%
%     \centering
%     \includegraphics[width=0.48\columnwidth]{results/hopper_reset.png}
%     % \caption{}
%     \label{fig:hopper_a}
% %\end{subfigure}
% \hfill
% % \begin{subfigure}[b]{0.235\textwidth}
% \subfloat[]{%
%     \centering
%     \includegraphics[width=0.48\columnwidth]{results/cheetah_reset.png}
%     % \caption{}
%     \label{fig:half_cheetah_b}
% } 
% % \end{subfigure}
% \caption{Legged systems tested: (a) Hopper. (b) Half Cheetah}
% \label{fig:main_figure}
% \end{figure}

\subsubsection{Hopper}
\begin{figure}[t]
    \centering
    \includegraphics[width=0.8\columnwidth]{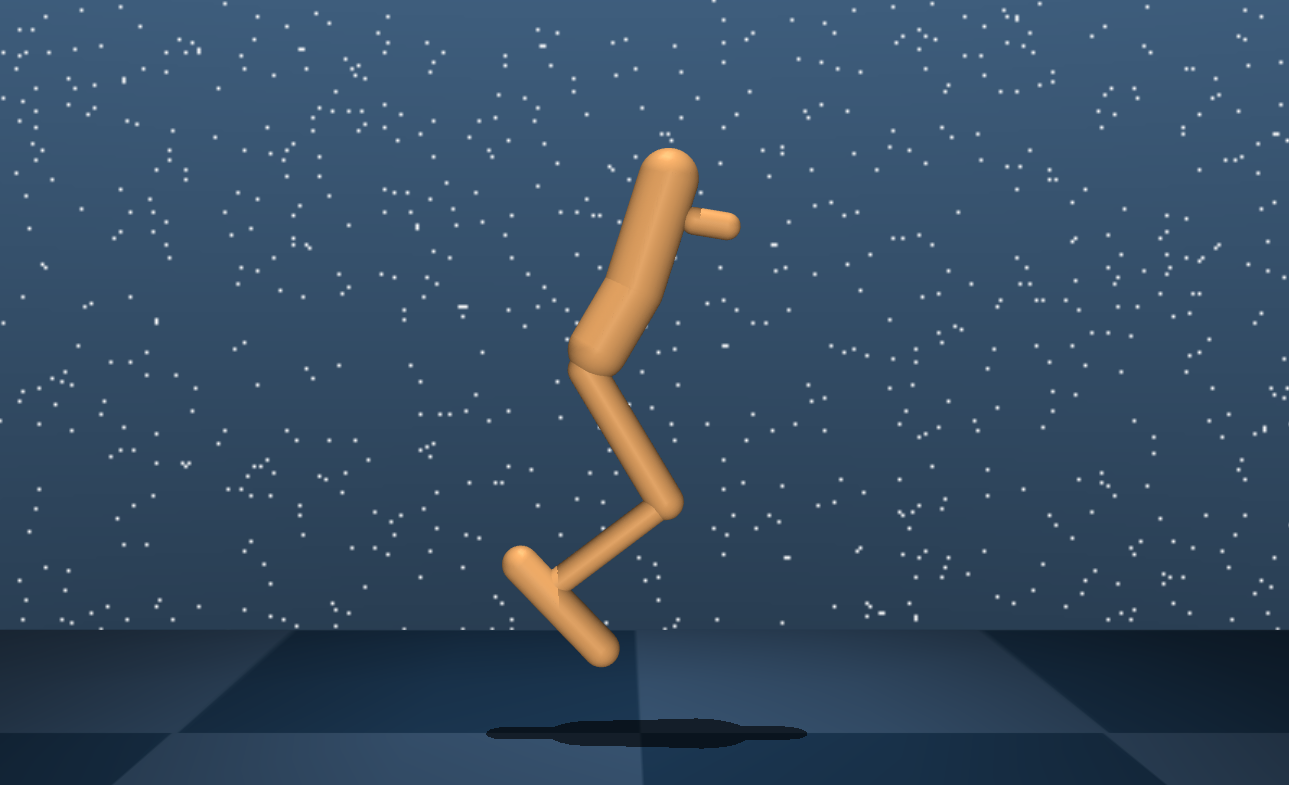}
    \caption{Snapshot of the {Hopper} legged system.}
    \label{fig:hopper}
\end{figure}

\begin{figure}
    \centering
    \includegraphics[width=\columnwidth]{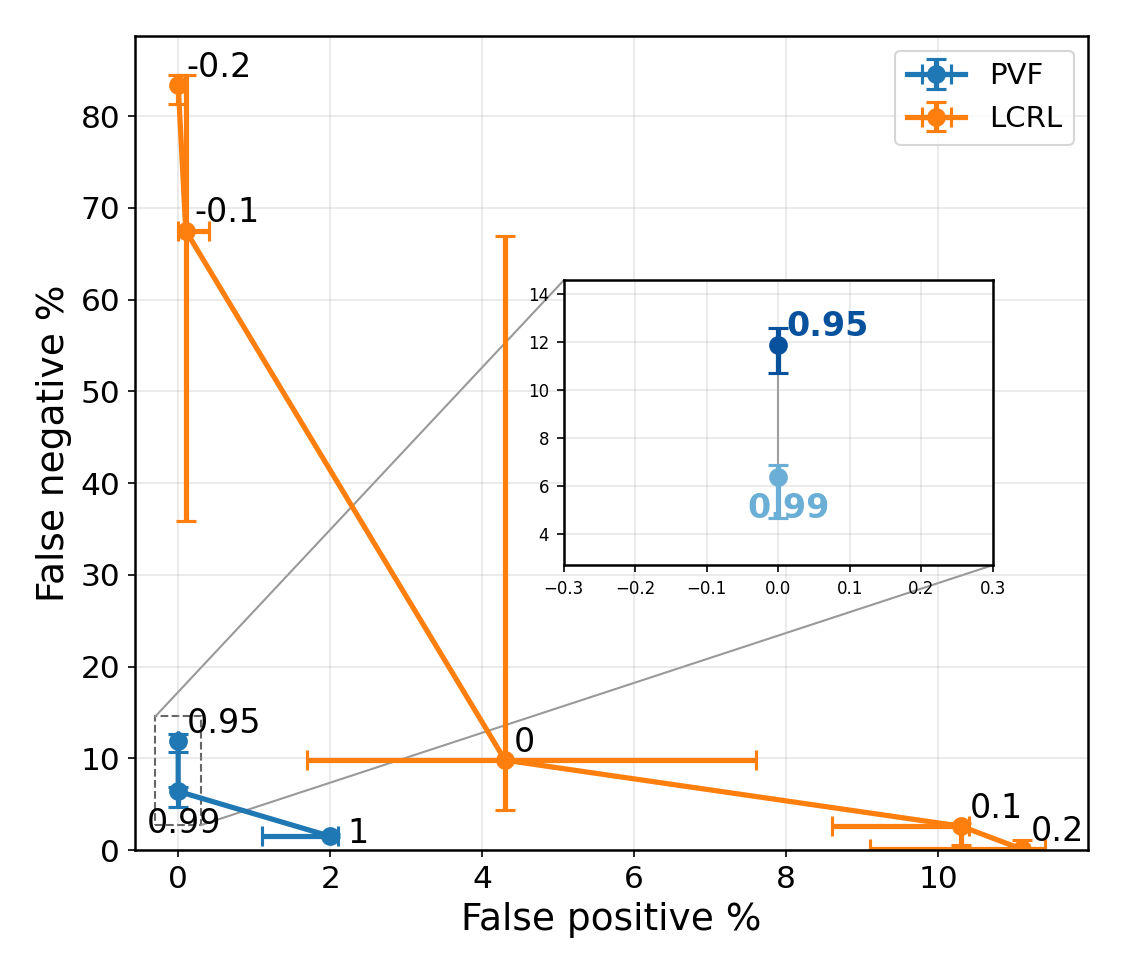}
    \caption{{Comparison between PVF and LCRL in terms of accuracy, for Hopper experiment.} Thresholds for fail detection reported near the medians. }
    \label{fig:hopper_res}
\end{figure}

\label{subsec:hopper}
This test involves the common RL benchmarking system Hopper~\cite{Durrant-White2012}, shown in Fig.~\ref{fig:hopper}, characterized by non-smooth contact dynamics and a 12-dimensional state. 
The robot started at a random state, with the horizontal position $x_h \in [-10,60]$ m, and has to reach $x_h\ge 50$ m without falling, while subjected to disturbances in horizontal and vertical velocity. 
The velocity disturbances are modeled as a zero-mean Gaussian truncated at $4 \sigma = 0.15$ m/s.
The comparison between PVF and LCRL is reported in Fig.~\ref{fig:hopper_res}. 
Even in this case PVF was more accurate on average than LCRL both in terms of FN and FP at the nominal thresholds. 
Under nominal conditions, the median FN rate for LCRL was about 
$4\%$ higher than for PVF, although the minimum FN rates across runs were very close, whereas PVF performed markedly better in terms of FP, achieving nearly $0\%$.
This could be due to the fact that the disturbance policy learned by LCRL was not truly worst-case, and that PVF encountered more adversarial disturbances during data collection.

% \subsubsection{Half Cheetah}
% This test focuses on a larger system, the Half Cheetah \cite{Wawrzyński2009}, shown in Fig.~\ref{fig:half_cheetah_b}, featuring an 18-dimensional state. 
% The comparison is reported in Fig.~\ref{fig:half_cheetah}. 
% The velocity disturbances are modeled as a Gaussian truncated at $4 \sigma = 0.30$ m/s, the target set is $x_h\ge90$ and the episode starts randomly with $x_h \in [-10,100]$ . 
% LCRL performed better in terms of FN at the nominal condition, but at the price of FP errors above $10\%$. 
% The large errors in this case for PVF are most likely due to the characteristics of the task: being the Half Cheetah intrinsically more robust to disturbances because of its large support area, PVF could have not been able to observe a sufficiently high number of failures to accurately estimate the failure probability.

\subsection{Computation times}
Table~\ref{tab:computation-time} reports the mean computation time for each method. 
In our implementation, PVF required more computation, primarily because of its larger transition dataset and repeated stochastic rollouts.
% PVF is consistently more expensive than LCRL, as it considers the full distribution of the noise rather than only its worst-case realization, which requires substantially more data and, consequently, longer data collection and training times. 
For Hopper, LCRL was run with a fixed policy, which explains its comparatively low computation time on this system w.r.t. to Drones racing; nonetheless, allowing LCRL additional computation time did not lead to any improvement in performance. 
For PVF, the size of the generated dataset is the main driver of computation time. 
We believe that the method could potentially work well even with smaller datasets, yielding significantly shorter data collection and training times than those reported in Table~\ref{tab:computation-time}, and that implementing a technique that concentrates the training where the learned probability error relative to the targets is larger could further lower the times. However, more work is needed to validate such hypothesis.   
\begin{table}[tbp]
\centering
\caption{Mean computation time across the different runs.}
\label{tab:computation-time}
\resizebox{\columnwidth}{!}{%
\begin{tabular}{lcc}
\toprule
\sc{} & \sc{Drones Racing} & \sc{Hopper} \\
\midrule
LCRL & 3175 s & 361 s\\
PVF data collection & 677 s & 7145 s\\
PVF training & 9367 s & 23502 s\\
PVF total & 10044 s & 30647 s \\
\bottomrule
\end{tabular}
}
\end{table}

% \begin{figure}[tbp]
%     \centering
%     \includegraphics[width=\columnwidth]{results/half_cheetah_not_biased_fn_fp.png}
%     \caption{Comparison between PVF and LCRL in terms of accuracy, for Half Cheetah experiment.Thresholds for fail detection reported near the medians.}
%     \label{fig:half_cheetah}
% \end{figure}

\subsection{Ablation: undiscounted avoid-only formulation}
To validate the importance of the target set (as discussed in Section~\ref{ssec:comparison}), even in cases where our main interest is avoiding the constraint set, we train a PVF on Hopper, considering only failures.
This approach is equivalent to SQRL~\cite{Srinivasan2020} without discounting. 
If Hopper falls within $500$ steps the episode fails.
We computed the TD target as:
\begin{equation}
    V_{target}(x, x') :=
    \begin{cases}
        0 & \text{if } x \in \mathcal{C} \\
        V(x') & \text{otherwise}
    \end{cases}
\end{equation}
We validated the accuracy of the learned value function computing the Wilson confidence intervals, as in the previous tests. 
The classification threshold was set to $0.99$, corresponding to a $1\%$ failure probability.

\begin{table}[tbp]
    \centering
    \caption{Failures-only Network results for Hopper.}
    \label{tab:failures-only}
    \begin{tabular}{lc}
        \toprule
        \sc{Metric} & \sc{Value (\%)} \\
        \midrule
        {True Positive}  & 74.6 \\
        {False Positive} & 23.6 \\
        {True Negative}  & 0.0  \\
        {False Negative} & 0.0  \\
        {Unclear Cases}  & 1.8  \\
        \bottomrule
    \end{tabular}
\end{table}

The results reported in Table~\ref{tab:failures-only} show that this approach was not able to detect the safe states, because the network always classified the states as unsafe, acting extremely cautiously. 
Intuitively, this can be explained by the fact that, since the training is self-supervised, except for the failure signals, the network is only forced to go towards the target associated to failures.

\subsection{Application: model predictive shielding}
As an example of a potential application, we have used the learned PVF and the associated (backup) policy as a shield for another policy~\cite{Bastani2021}. 
By selecting a desired probability level, the sublevel-set of $V(\cdot)$ approximates the associated probabilistic reach-avoid set.
Testing membership to this set, a so-called safety filter can decide when to trigger the switch from a nominal policy to the backup policy. 
The settings of the reach-avoid task are the same as in Section~\ref{subsec:hopper}.
For this test we use two policies:
\begin{itemize}
    \item the \emph{nominal policy}, trained on Hopper without disturbances, with a reward focused on forward speed;
    \item the \emph{backup policy}, trained on Hopper with disturbances, using a reward that assigns a lower weight to forward speed; among the checkpoints saved during RL training, we retain the one with the highest mean episode length. The probabilistic reach-avoid set is computed using this policy.
\end{itemize}
Starting from the same initial state, whose estimated probability of success exceeds $0.99$, the Hopper's task is to move forward $6$ m without falling. 
We performed $100$ rollouts of the nominal policy and, under the same disturbance realizations, $100$ rollouts of the nominal policy shielded by the backup policy. 
The shield triggers whenever the mean predicted probability of success at the next state falls below $0.99$. 
To estimate this probability, we query the simulator $n$ times from the current state and average the PVF estimate over the resulting $n$ successor states. 
Three noise levels were tested, even though the used PVF network was trained only on rollouts subject to zero-mean Gaussian noise truncated at $4\sigma$, with a maximum disturbance of $4\sigma = 0.15$.
The test was repeated for $11$ different seeds, using $n=100$ with identical disturbance realizations applied to both the shielded and unshielded cases. 
Mean and standard deviation of the success rate across all runs are reported in Table \ref{tab:shielding_results}.
With shielding active, the number of failures was consistently lower than in the unshielded case, even for noise levels exceeding those used to compute the PVF. 
At the disturbance bound $4\sigma = 0.15$, the shielded execution achieved, on average, $6.4\%$ more successes than the unshielded policy, with roughly half its standard deviation. 
At the higher noise bound $4\sigma = 0.20$, the gap between the two methods widened, with shielding achieving $7.9\%$ more successes on average. 
We believe that the gap between the desired success probability (99\%) and the lower observed success rate is due to the distribution mismatch between the training-set states and the states visited by the nominal policy at deployment.
\begin{table}[t]
\centering
\caption{Mean $\pm$ standard deviation of success counts (out of 100 trials) across 11 seeds, by noise level.}
\label{tab:shielding_results}
\begin{tabular}{lccc}
\toprule
 & \multicolumn{3}{c}{\sc{Noise Level} ($4\sigma$)} \\
\cmidrule{2-4}
 & $0.10$ & $0.15$ & $0.20$ \\
\midrule
\sc{Shielded}   & $99.1 \pm 1.6$  & $97.2 \pm 4.2$ & $91.0 \pm 8.3$ \\
\sc{Unshielded} & $97.1 \pm 5.2$  & $90.8 \pm 7.7$ & $83.1 \pm 10.0$ \\
\bottomrule
\end{tabular}
\end{table}

\section{Conclusions}
In this article we presented PVF, a novel method for computing probabilistic reach-avoid sets. We compared it against LCRL, a recent and promising HJ-reachability method for computing robust reach-avoid sets. Under random disturbances, PVF achieved better classification accuracy when evaluated against a ground truth certified via Wilson confidence intervals at a high confidence level.
Despite these encouraging results, the proposed approach still presents some limitations that would require further attention. 
Most importantly, our method currently assumes that a policy is available for controlling the system. 
We plan to extend our method to jointly learn the probabilistic value function and a policy to maximize the associated success probability.
Another critical point is computational efficiency, which is currently a weakness of the method. 
In this respect we want to investigate ways to intensify training on state-space regions where the probability estimate is more uncertain. 
Finally, we plan to apply and validate our method on larger systems, such as quadruped robots navigating obstacle-rich environments, and investigate the use of PVF for shielding in safety critical applications.
\addtolength{\textheight}{-0cm}   % This command serves to balance the column lengths

%%%%%%%%%%%%%%%%%%%%%%%%%%%%%%%%%%%%%%%%%%%%%%%%%%%%%%%%%%%%%%%%%%%%%%%%%%%%%%%%

% \newpage
\bibliographystyle{IEEEtran}
\bibliography{IEEEabrv,references}

@misc{lillicrap2019,
      title={Continuous control with deep reinforcement learning}, 
      author={Timothy P. Lillicrap and Jonathan J. Hunt and Alexander Pritzel and Nicolas Heess and Tom Erez and Yuval Tassa and David Silver and Daan Wierstra},
      year={2019},
      eprint={1509.02971},
      archivePrefix={arXiv},
      primaryClass={cs.LG},
      url={https://arxiv.org/abs/1509.02971}, 
}

@inproceedings{Bastani2021,
author = {Bastani, Osbert and Li, Shuo and Xu, Anton},
booktitle = {Robotics Science and Systems},
title = {{Safe Reinforcement Learning via Statistical Model Predictive Shielding}},
year = {2021}
}

@article{LaRocca2023,
author = {{La Rocca}, Asia and Saveriano, Matteo and {Del Prete}, Andrea},
title = {{VBOC: Learning the Viability Boundary of a Robot Manipulator using Optimal Control}},
journal = {IEEE Robotics and Automation Letters},
year = {2023}
}

@article{Blanchini1999,
   author = {F. Blanchini},
   doi = {10.1016/S0005-1098(99)00113-2},
   issn = {00051098},
   number = {11},
   journal = {Automatica},
   title = {Set invariance in control},
   volume = {35},
   year = {1999},
   pages = {1747--1767}
}

@inproceedings{Bansal2017,
   author = {Somil Bansal and Mo Chen and Sylvia Herbert and Claire J. Tomlin},
   doi = {10.1109/CDC.2017.8263977},
   isbn = {9781509028733},
   booktitle = {IEEE Conference on Decision and Control (CDC)},
   title = {Hamilton-jacobi reachability: A brief overview and recent advances},
   year = {2017},
}

@book{Grune2017,
  title={Nonlinear model predictive control},
  author={Gr{\"u}ne, Lars and Pannek, J{\"u}rgen},
  year={2017},
  publisher={Springer}
}

@INPROCEEDINGS{Bansal2020,
  author={Bansal, Somil and Tomlin, Claire J.},
  booktitle={IEEE International Conference on Robotics and Automation}, 
  title={DeepReach: A Deep Learning Approach to High-Dimensional Reachability}, 
  year={2021},
  volume={},
  number={},
  xpages={1817-1824},
  doi={10.1109/ICRA48506.2021.9561949}}

@INPROCEEDINGS{Fisac2019,
  author={Fisac, Jaime F. and Lugovoy, Neil F. and Rubies-Royo, Vicenç and Ghosh, Shromona and Tomlin, Claire J.},
  booktitle={IEEE International Conference on Robotics and Automation}, 
  title={Bridging Hamilton-Jacobi Safety Analysis and Reinforcement Learning}, 
  year={2019},
  volume={},
  number={},
  pages={8550-8556},
  doi={10.1109/ICRA.2019.8794107}}

@ARTICLE{Li2025,
  author={Li, Jingqi and Lee, Donggun and Lee, Jaewon and Dong, Kris Shengjun and Sojoudi, Somayeh and Tomlin, Claire},
  journal={IEEE Robotics and Automation Letters (RA-L)}, 
  title={Certifiable Reachability Learning Using a New Lipschitz Continuous Value Function}, 
  year={2025},
  volume={10},
  number={4},
  xpages={3582-3589},
  doi={10.1109/LRA.2025.3535183}}

@misc{Srinivasan2020,
      title={Learning to be Safe: Deep RL with a Safety Critic}, 
      author={Krishnan Srinivasan and Benjamin Eysenbach and Sehoon Ha and Jie Tan and Chelsea Finn},
      year={2020},
      xeprint={2010.14603},
      xarchivePrefix={arXiv},
      primaryClass={cs.LG},
      xurl={https://arxiv.org/abs/2010.14603}, 
      howpublished = {arXiv:2010.14603},
}

@article{Summers2010,
title = {Verification of discrete time stochastic hybrid systems: A stochastic reach-avoid decision problem},
journal = {Automatica},
volume = {46},
number = {12},
xpages = {1951--1961},
year = {2010},
author = {Sean Summers and John Lygeros}
}

@article{Abate2008,
title = {Probabilistic reachability and safety for controlled discrete time stochastic hybrid systems},
journal = {Automatica},
volume = {44},
number = {11},
xpages = {2724-2734},
year = {2008},
author = {Alessandro Abate and Maria Prandini and John Lygeros and Shankar Sastry}
}

@article{brown2001interval,
  author  = {Brown, Lawrence D. and Cai, T. Tony and DasGupta, Anirban},
  title   = {Interval Estimation for a Binomial Proportion},
  journal = {Statistical Science},
  volume  = {16},
  number  = {2},
  xpages   = {101--133},
  year    = {2001},
  doi     = {10.1214/ss/1009213286}
}

@INPROCEEDINGS{Durrant-White2012,
  author={Durrant-Whyte, Hugh and Roy, Nicholas and Abbeel, Pieter},
  booktitle={Robotics: Science and Systems VII}, 
  title={Infinite-Horizon Model Predictive Control for Periodic Tasks with Contacts}, 
  year={2012},
  volume={},
  number={},
  xpages={73-80},
  doi={},
  }

@article{td_sutton_1988,
  title = {Learning to Predict by the Methods of Temporal Differences},
  author = {Sutton, Richard S.},
  year = {1988},
  xmonth = aug,
  journal = {Machine Learning},
  volume = {3},
  number = {1},
  xpages = {9--44},
  issn = {0885-6125, 1573-0565},
  doi = {10.1007/BF00115009},
  urldate = {2025-09-04},
}

@misc{mujoco_playground_2025,
  title = {MuJoCo Playground: An open-source framework for GPU-accelerated robot learning and sim-to-real transfer.},
  author = {Zakka, Kevin and Tabanpour, Baruch and Liao, Qiayuan and Haiderbhai, Mustafa and Holt, Samuel and Luo, Jing Yuan and Allshire, Arthur and Frey, Erik and Sreenath, Koushil and Kahrs, Lueder A. and Sferrazza, Carlo and Tassa, Yuval and Abbeel, Pieter},
  year = {2025},
  publisher = {GitHub},
  url = {https://github.com/google-deepmind/mujoco_playground}
}

@misc{schulman2017ppo,
      title={Proximal Policy Optimization Algorithms}, 
      author={John Schulman and Filip Wolski and Prafulla Dhariwal and Alec Radford and Oleg Klimov},
      year={2017},
      xeprint={1707.06347},
      xarchivePrefix={arXiv},
      primaryClass={cs.LG},
      xurl={https://arxiv.org/abs/1707.06347}, 
      howpublished = {arXiv:1707.06347},
}

\end{document}